\documentclass[lettersize,journal]{IEEEtran}
\usepackage{amsmath,amsfonts}
\usepackage{algorithmic}
\usepackage{algorithm}
\usepackage{array}
\usepackage[caption=false,font=normalsize,labelfont=sf,textfont=sf]{subfig}
\usepackage{textcomp}
\usepackage{stfloats}
\usepackage{url}
\usepackage{verbatim}
\usepackage{graphicx}
\usepackage{cite}
\usepackage[normalem]{ulem}
\usepackage{booktabs}
\usepackage{tablefootnote}
\usepackage{makecell}
\usepackage{multirow}
\usepackage{xcolor,colortbl}
\definecolor{Tabcolor}{rgb}{0.99, 0.96, 0.94}
\newcolumntype{a}{>{\columncolor{Tabcolor}}c}
\newcommand{\g}[1]{\textcolor{gray}{#1}}
\def\etal{\emph{et al}.}

\begin{document}

\title{Exploring Multi-Modal Contextual Knowledge for Open-Vocabulary Object Detection}
\author{Yifan Xu$^*$$^\dagger$, Mengdan Zhang$^\dagger$, Xiaoshan Yang, Changsheng Xu,~\IEEEmembership{Fellow,~IEEE}
\thanks{Manuscript received September 1, 2023.}
\thanks{Corresponding author: Changsheng Xu.}
\thanks{$^*$Work partially done during internship at Tencent Youtu Lab. $^\dagger$Equal contribution.}
\thanks{Yifan Xu, Xiaoshan Yang, and Changsheng Xu are with MAIS, Institute of Automation, Chinese Academy of Sciences, University of the Chinese Academy of Sciences, and Peng Cheng Laboratory. E-mail: \{yifan.xu, xiaoshan.yang, csxu\}@nlpr.ia.ac.cn

Mengdan Zhang is with Tencent Youtu Lab. E-mail: davinazhang@tencent.com}}

\markboth{IEEE Transactions on Image Processing,~Vol.~14, No.~8, August~2021}%
{Shell \MakeLowercase{\textit{et al.}}: A Sample Article Using IEEEtran.cls for IEEE Journals}

\IEEEpubid{0000--0000/00\$00.00~\copyright~2021 IEEE}

\maketitle

\begin{abstract}
In this paper, we for the first time explore helpful multi-modal contextual knowledge to understand novel categories for open-vocabulary object detection (OVD). The multi-modal contextual knowledge stands for the  joint relationship across regions and words. However, it is challenging to incorporate such multi-modal contextual knowledge into OVD. The reason is that previous detection frameworks fail to jointly model multi-modal contextual knowledge, as object detectors only support vision inputs and no caption description is provided at test time.
To this end, we propose a multi-modal contextual knowledge distillation framework, \textbf{MMC-Det}, to transfer the learned contextual knowledge from a teacher fusion transformer with diverse multi-modal masked language modeling (D-MLM) to a student detector. The diverse multi-modal masked language modeling is realized by an object divergence constraint upon traditional multi-modal masked language modeling (MLM), in order to extract fine-grained region-level visual contexts, which are vital to object detection.
Extensive experiments performed upon various detection datasets show the effectiveness of our multi-modal context learning strategy, where our approach well outperforms the recent state-of-the-art methods.
\end{abstract}


\section{Introduction}
\label{sec:intro}
\begin{figure}[t]
    \centering
    \includegraphics[width=0.45\textwidth]{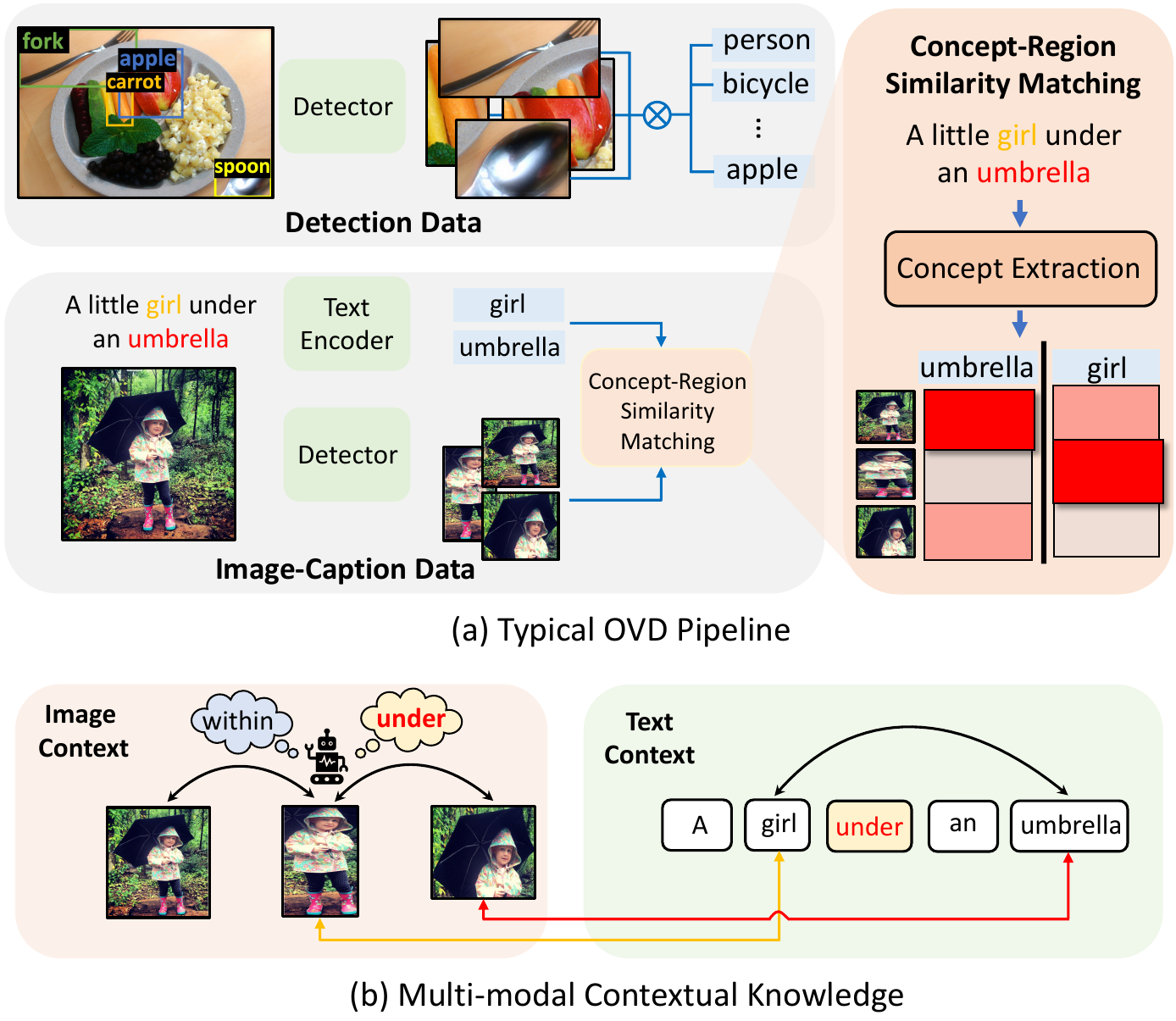}
  \caption{
  (a) A typical OVD pipeline utilizes both detection and image-caption data. The localization branch is omitted for concision. For detection data, the training process is similar to the one in traditional detection except the classifier with category text embeddings. For image-caption data, the region proposals and concepts in a caption are first respectively extracted via a detector and a text encoder, then aligned through concept-region dot-product similarity matching.
  (b) The multi-modal contextual knowledge stands for the joint relationship across
regions and words,
  }
    \label{fig:fig1}
\end{figure}

\IEEEPARstart{O}{pen}-vocabulary object detection (OVD)~\cite{zsd} aims to detect open-world instances beyond base categories that have well-labeled bounding boxes for training. To achieve this goal, a wide range of category vocabularies is required, which is commonly acquired by large-scale detection annotations in traditional object detection. On account of costly annotations for large-category detection datasets, an intuitive idea shared by OVD methods is to enlarge the category vocabularies via massive image-caption pairs~\cite{cc3m,coco-caption}, which are readily available via internet sources. Fig.\,\ref{fig:fig1}(a) shows a typical OVD pipeline. Different from image-level recognition~\cite{clip,align}, the main challenge of OVD is to seek region-level supervision~\cite{detic,regionclip} from image-caption pairs. Recently, there are some successful attempts~\cite{regionclip,vild,clip,detic,detclip} to address the challenge of open-vocabulary object detection. For example, OVR-CNN~\cite{ovrcnn} pretrained the detector on image-text pairs using contrastive learning. RegionCLIP~\cite{regionclip} finetuned image-level pretrained model CLIP~\cite{clip} with pseudo region-text pairs from massive image-text data. Detic~\cite{detic} assigned all concept words (typically nouns) in the captions to one max-size region proposal. 

\IEEEpubidadjcol

Although these methods have achieved significant progress in open-vocabulary object detection, they separately model the image and text contexts and directly align them via simple concept-region similarity matching, as shown in Fig.\,\ref{fig:fig1}(a). This kind of strategy will limit the open-vocabulary performances because they neglect a joint contextual relationship across the images and texts. Take Fig.\,\ref{fig:fig1}(a) as an example, a typical OVD pipeline separately matches the concepts ``umbrella'' and ``girl'' with the regions. This causes the model to falsely match the novel class ``umbrella'' with a coarse region, as the model does not actually know how the novel concept ``umbrella'' looks like. As shown in Fig.\,\ref{fig:fig1}(b), if jointly considering the location relationship of regions and the relationship of concepts (\emph{e.g.}, ``girl under an umbrella''), we can get additional clues that the region of ``girl'' is probably under the region of ``umbrella'', therefore acquiring more accurate region-concept alignment. 
We denote such helpful information as multi-modal contextual knowledge, namely, the joint contextual relationship across the images and texts. By utilizing multi-modal contextual knowledge, we can acquire additional clues for region-level supervision.

\begin{figure}[t]
    \centering
    \includegraphics[width=0.45\textwidth]{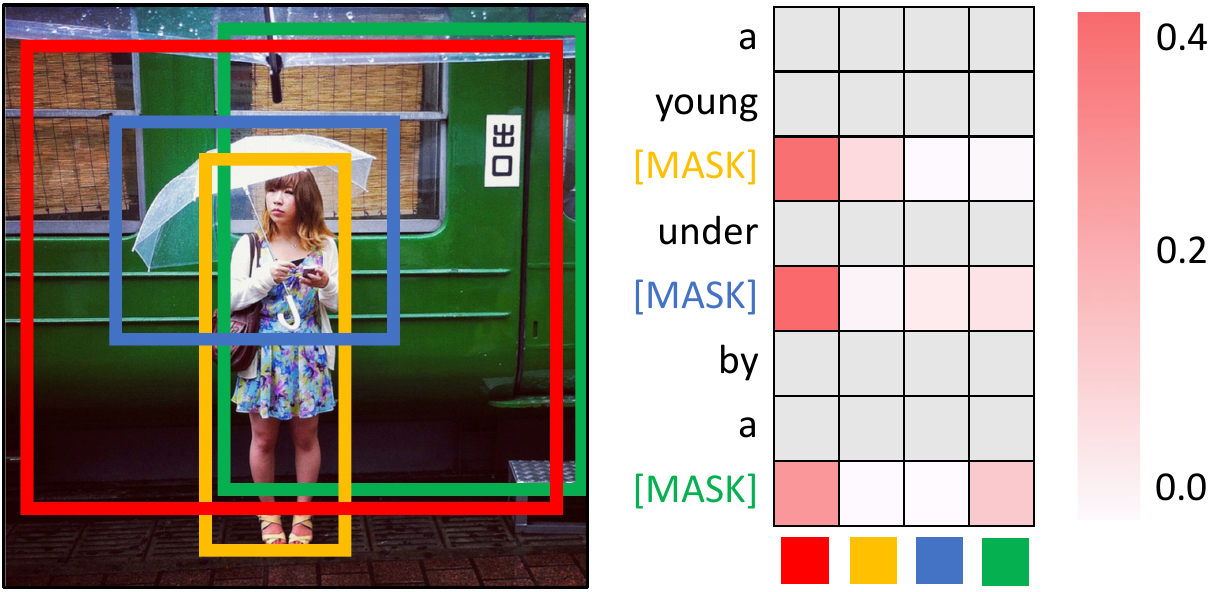}
  \caption{
  The attention scores of the masked concepts on the input regions in the last self-attention layer of the fusion transformer with vanilla MLM. All masked concepts highly activate on a similar global region.
  }
    \label{fig:vanilla_mlm}
\end{figure}

To this end, we explore multi-modal contextual knowledge to help open-vocabulary object detection. Multi-modal masked language modeling~\cite{albef,vilt} (MLM) has been demonstrated to be an effective context understanding approach in modern multi-modal pretraining models that generates corresponding words after taking the image and the partially-masked text as the joint inputs of a dense-modeling fusion transformer~\cite{transformer}. The core idea of MLM~\cite{vilbert,bert,albef} is that if a masked concept word is correctly predicted, it should jointly model the contextual knowledge in both images and texts during forward inference, which is reflected by corresponding attention activation~\cite{visualbert}. The attention mechanism is proved to be able to implicitly model the contextual knowledge and highly activate on related tokens~\cite{evovit,visualbert}. In light of this, a natural idea springs up, \emph{i.e.}, utilizing the contextual knowledge learned by MLM to help open-vocabulary detection. Nevertheless, how to incorporate such contextual knowledge into OVD faces two challenges. 
1) It is difficult to jointly model the contextual knowledge in images and texts in previous detection frameworks, as detectors only support vision inputs and no caption description is provided at test time. 2) Vanilla MLM is not sensitive to region-level information that is vital to detection despite its superiority in modeling multi-modal contextual knowledge. As shown in Fig.\,\ref{fig:vanilla_mlm}, most masked words highly activate their attention on a global region in the self-attention layers. This indicates that vanilla MLM may cheaply integrate coarse visual contextual information for context alignment. 

This work proposes a novel training framework \textbf{MMC-Det} to address the above two challenges, and incorporates the \textbf{m}ulti-\textbf{m}odal \textbf{c}ontextual knowledge into \textbf{det}ectors to improve the open-vocabulary detection performance. Illustated in Fig.\,\ref{fig:method}, MMC-Det is a multi-modal contextual knowledge distillation framework.
This framework transfers the contextual knowledge learned by a multi-modal fusion transformer to a student detector, thus addressing the first challenge. Specifically, we reuse the attention activation of the self-attention mechanism~\cite{transformer} within the fusion transformer as a soft label to supervise the concept-region similarity matching in the classifier of the detector. As for the second challenge, we introduce a diverse multi-modal masked language modeling (D-MLM) strategy. Concretely, an object divergence constraint upon MLM is proposed to increase the attention variance in the self-attention layers of the fusion transformer. This encourages each masked concept to activate diverse attention on its exclusive proposals, enabling the fusion transformer to focus on fine-grained region-level visual contexts.

In a nutshell, our main contributions can be summarized as:

\begin{itemize}
    \item We explore a novel idea that utilizes the multi-modal contextual knowledge to investigate additional localization clues for novel classes in open-vocabulary object detection (OVD).  To our knowledge, we are the first to explore multi-modal contextual knowledge in OVD.
    \item We propose a multi-modal contextual knowledge distillation framework MMC-Det to transfer the multi-modal contextual knowledge learned from a teacher fusion transformer with diverse multi-modal masked language modeling (D-MLM) to a student detector. The diverse multi-modal masked language modeling is detection-oriented and provides fine-grained region-level contextual knowledge.
    \item Extensive experiments on various detection datasets manifest substantial performance gains of our approach compared with the existing OVD methods.
\end{itemize}
\section{Related Works}
\label{sec: related_works}


\textbf{Open-vocabulary object detection~(OVD)}
aims to detect target instances beyond limited well-annotated base classes via vision-language supervision, which is commonly derived from massive image-caption pairs and pretrained vision-language models like CLIP~\cite{clip, medet}.
Different from grounded language-image pretraining~\cite{glip,mqdet,mdetr,xdetr,detclip}, which relies on numerous grounding and detection data for large-scale fine-grained pretraining, OVD utilizes limited detection data to generalize to open-world instances, and is potential to further benefit from larger grounding datasets~\cite{flickr,VG,GQA}.
Zareian~\etal\cite{ovrcnn} introduced OVD task for the first time and proposed OVR-CNN, where a visual encoder was first pretrained on image-caption pairs to learn rich vocabulary, and then fine-tuned on detection data with only base classes.
ViLD~\cite{vild} used pretrained CLIP~\cite{clip} to distill knowledge into two-stage object detectors~\cite{faster_rcnn}.
RegionCLIP~\cite{regionclip} extended CLIP to learn region-level visual representations, thus enabling fine-grained alignment between image regions and class concepts during fine-tuning.
Detic~\cite{detic} simultaneously learned object localization and fine-grained vision-language matching by using the max-size proposal to assign image-level labels.
Hanoona~\etal\cite{object_centric} used external class-agnostic detectors MViT~\cite{mvit} to generate region-concept pairs and learned fine-grained vision-language matching for object-centric OVD.
Despite impressive performance, most current OVD methods directly untilize the concept-region dot-product similarity to learn fine-grained vision-language matching, which ignore the joint contextual relationship within images and texts.
For a more generalized perspective, we pursue multi-modal contextual relationship learning in the detection framework.

\textbf{Multi-modal contextual representation learning} has been mainly explored by recent vision-language pretraining~(VLP) works~\cite{transformer_suervey, vilbert,su2019vl,li2020unicoder,yang2022vision,visualbert,albef,kwon2022masked,luo2022towards,yang2023membridge} and corresponding downstream tasks such as visual question answering~\cite{vqa,goyal2017making,zeng2022video,zhang2023reducing} and visual commonsense reasoning~\cite{vcr,li2023joint,zhao2022videoabc}. 
Typically, a multi-layer transformer~\cite{transformer} is adopted as the modality fusion module to model both visual and linguistic contents.
To learn cross-modal context, three types of tasks are commonly used, \emph{i.e.}, masked language modeling~(MLM)~\cite{bert}, masked image modeling (MIM)~\cite{mae,simmim}, and image-text matching~(ITM)~\cite{li2019visual}. 
The first two tasks jointly learn context-aware representations for input tokens based on linguistic and visual contents.
The last task learns to predict whether an image and a text describe each other. 
Specifically, VisualBERT~\cite{visualbert} proposed to learn joint contextualized representations of vision and language by MLM and ITM. 
Unicoder-VL~\cite{li2020unicoder} directly predicted the categories of masked proposals.
Our framework is similar to ALBEF~\cite{albef}, which proposed an align-before-fuse multi-modal framework, \emph{i.e.}, first conducting modality alignment via contrastive learning, then feeding the aligned image-text tokens into a fusion transformer for further contextual interaction via MLM and ITM.
Different from these VLP works that aim to incorporate contextual knowledge into a late-fusion transformer, we explore to utilize the contextual knowledge from the fusion transformer learned by the generative task to guide the discriminative learning of the preceding detector.

\textbf{Knowledge distillation} aims to transfer the knowledge from a well-learned teacher model to a student model to help the learning of the student model.
Knowledge distillation is first proposed in~\cite{kd}, which distills the knowledge from a large model to an efficient small model by minimizing the Kullback–Leibler divergence~\cite{kl} between their logits distribution. A direct implementation of knowledge distillation is model compression~\cite{sun2019patient,gao2022disco,korattikara2015bayesian,luo2016face,song2022spot,li2022ckdf,huang2022feature,ge2022learning}. Our work is more related to another pipeline, namely privileged information distillation~\cite{lopez2015unifying,tu2022general}. The privileged information is available during training but not accessible during testing. Lopez-Paz \etal~\cite{lopez2015unifying} first propose to generalize distillation to incorporate privileged information. Gupta \etal~\cite{gupta2016cross} treat the extra modality as the privileged information for cross-modal distillation. Zhou \etal~\cite{zhou2020more} propose to use image-text matching model to distill word-region alignment information for image captioning. Recently, Gu \etal~\cite{vild} propose ViLD for open-vocabulary object detection to distill the open-vocabulary knowledge from a image-level pretrained model CLIP~\cite{clip} to a detector. 
Differently, our method distills the contextual knowledge from a generative multi-modal fusion transformer to a discriminative single-modal detector.



\section{Methodology}
\label{sec:method}

\subsection{Preliminary}
\textbf{Open-vocabulary object detection~(OVD).}
Given an image $I \in \mathbb{R}^{c\times h\times w}$ ($c$ = $3$ usually), open-vocabulary detection~\cite{detic,ovrcnn} solves two sub-problems: (1) localization: find all objects with their locations, each represented as a box $\mathbf{b}_{j} \in \mathbf{R}^{4}$ and a region feature $r_j \in \mathbf{R}^{d}$, and (2) 
classification: assign a category label $y_{j} \in Y_\textrm{test}= Y_\textrm{base} \cup Y_\textrm{novel} \cup bg$ to the $j$-th object.
Here $Y_\textrm{test}$ is the set of category vocabularies provided at test time. $Y_\textrm{base}$ is the category set available in training phase. $Y_\textrm{novel}$ is consisted of novel classes that are only available during testing, where $Y_\textrm{novel} \cap Y_\textrm{base} = \emptyset$. 
$bg$ denotes background. Commonly, the classification is conducted via dot-product similarity matching, namely,
\begin{equation}
    y_{j} = \mathop{\arg\max}_{y \in Y_\textrm{test}} \langle r_j, \Phi(y)\rangle,
\end{equation}
where $\langle \cdot, \cdot \rangle$ is the dot product of two vectors, $\Phi$ denotes a text encoder. In the following parts, we omit the text encoders in the dot product $\langle \cdot, \cdot \rangle$ for concision.

During training, a detection dataset $D^\textrm{det}=\{(I, \{y, \mathbf{b}\})_{i}\}_{i=1}^{|D^\textrm{det}|}$ with both categories and boxes is given, where $y \in Y_\textrm{base}$.
Besides the detection dataset, an image-caption dataset $D^\textrm{cap}=\{(I, T)_{i}\}_{i=1}^{|D^\textrm{cap}|}$ is also provided, where text $T$ is the corresponding caption that describes image $I$ and implicitly contains a wide range of category concepts including $Y_\textrm{test}$.

A baseline OVD framework contains two main types of losses, namely, the detection loss $\mathcal{L}_\textrm{det}$ and the caption loss $\mathcal{L}_\textrm{cap}$.
The detection loss is the same as that in traditional object detection~\cite{faster_rcnn,uiu}, which is only applied on detection data, as:
\begin{equation}
    \mathcal{L}_\textrm{det}(D^\textrm{det})=\mathcal{L}_\textrm{cls} + \mathcal{L}_\textrm{reg} + \mathcal{L}_\textrm{rpn},
    \label{eqn: loss_det}
\end{equation}
where $\mathcal{L}_\textrm{cls}$, $\mathcal{L}_\textrm{reg}$, and $\mathcal{L}_\textrm{rpn}$ represent the classification loss, localization regression loss, and region proposal loss, respectively.
The caption loss only takes image-caption pairs as inputs, and is commonly realized by a certain variant of image-level contrastive learning $\mathcal{L}_\textrm{con}$~\cite{clip, ovrcnn, regionclip}.
In this paper, we utilize 
grounding contrastive function~\cite{ovrcnn} as our baseline caption loss with slight modification, \emph{i.e.}, replacing the image grids with region proposals.
Specifically, given a batch of image-caption pairs $\{\mathcal{B}_I, \mathcal{B}_T\}$ from $D^\textrm{cap}$, where $\mathcal{B}_I$ and $\mathcal{B}_T$ are the images and the corresponding captions batches (usually $|\mathcal{B}_I|=|\mathcal{B}_T|$), respectively. For each image $I$ and its caption $T$, we first extract region proposal features $R^{I}=\{r_j\}_{j=1}^{n_{I}}$ in $I$ and every word token $W^{T}=\{w_i\}_{i=1}^{n_{T}}$ in $T$ through a detector~\cite{faster_rcnn,centernet} and a text encoder~\cite{bert,clip}, respectively. 
The weighted average of local grounding scores for word-region pairs $\langle W^{T},R^{I}\rangle_S$ is then computed:
\begin{equation}
\langle W^{T}, R^{I}\rangle_S=\frac{1}{n_{T}} \sum_{i=1}^{n_{T}} \sum_{j=1}^{n_{I}} a_{i, j}\left\langle w_i, r_j\right\rangle,
\label{eqn:ground_score}
\end{equation}
where $n_{I}$ and $n_{T}$ are the number of proposals and word tokens, and 
\begin{equation}
a_{i, j}=\frac{\exp \left\langle w_i, r_j\right\rangle}{\sum_{j^{\prime}=1}^{n_{I}} \exp \left\langle w_i, r_{j^{\prime}} \right\rangle}.
\label{eqn:sim_weight}
\end{equation}
The image-text grounding objective is then defined as:
\begin{equation}
\mathcal{L}_\textrm{con}(I, T)=-\log \frac{\exp \langle W^{T}, R^{I}\rangle_S}{\sum_{T^{\prime} \in \mathcal{B}_T} \exp \left\langle W^{T^{\prime}}, R^{I}\right\rangle_S},
\label{eqn:loss_contrast}
\end{equation}
and
\begin{equation}
\mathcal{L}_\textrm{con}(T, I)=-\log \frac{\exp \langle W^{T}, R^{I}\rangle_S}{\sum_{I^{\prime} \in \mathcal{B}_I} \exp \left\langle W^{T}, R^{I^{\prime}}\right\rangle_S}.
\label{eqn:loss_contrast_text}
\end{equation}
Accordingly, the final caption loss is:
\begin{equation}
\mathcal{L}_\textrm{cap}(I,T)=\mathcal{L}_\textrm{con}(I,T)+\mathcal{L}_\textrm{con}(T,I).
\label{eqn:loss_cap}
\end{equation}
In addition to the caption loss in Eqn.\,(\ref{eqn:loss_cap}), an additional image-level pseudo labeling loss~\cite{detic} is commonly used for better training stability. 
Specifically, we first extract object concepts $C^{T}=\{c_{i}\}$ from captions $T$, e.g., $T=$ ``a man beside a dog'' and $C^{T}=\{\textrm{``man''}, \textrm{``dog''}\}$. Then, we represent the full image as an additional proposal $r_{g}$, and conduct multi-label classification based on the target concept categories $C^{T}$, formulated as:

\begin{equation}
    \mathcal{L}_\textrm{img}= \frac{1}{|C^{T}|} \sum_{c_{i} \in C^{T}} \mathcal{L}_\textrm{CE}(f(r_{g}), c_{i}),
    \label{eqn:loss_image}
\end{equation}
where $f$ is a classification head in the detector and $f(r_{g})$ denotes the predicted logits. $\mathcal{L}_\mathrm{CE}$ is typically a cross entropy loss. To this end, a baseline open-vocabulary object detector is trained through:
\begin{equation} \mathcal{L}_\mathrm{ovd}=\mathcal{L}_\textrm{det}+\mathcal{L}_\textrm{cap}+\mathcal{L}_\mathrm{img},
\label{eqn:loss_baseline}
\end{equation}
where $\mathcal{L}_\textrm{det}$, $\mathcal{L}_\textrm{cap}$, and $\mathcal{L}_\textrm{img}$ are the detection loss in Eqn.\,(\ref{eqn: loss_det}), the caption loss in Eqn.\,(\ref{eqn:loss_cap}), and the image loss in Eqn.\,(\ref{eqn:loss_image}).

\begin{figure*}[t]
    \centering
    \includegraphics[width=\textwidth]{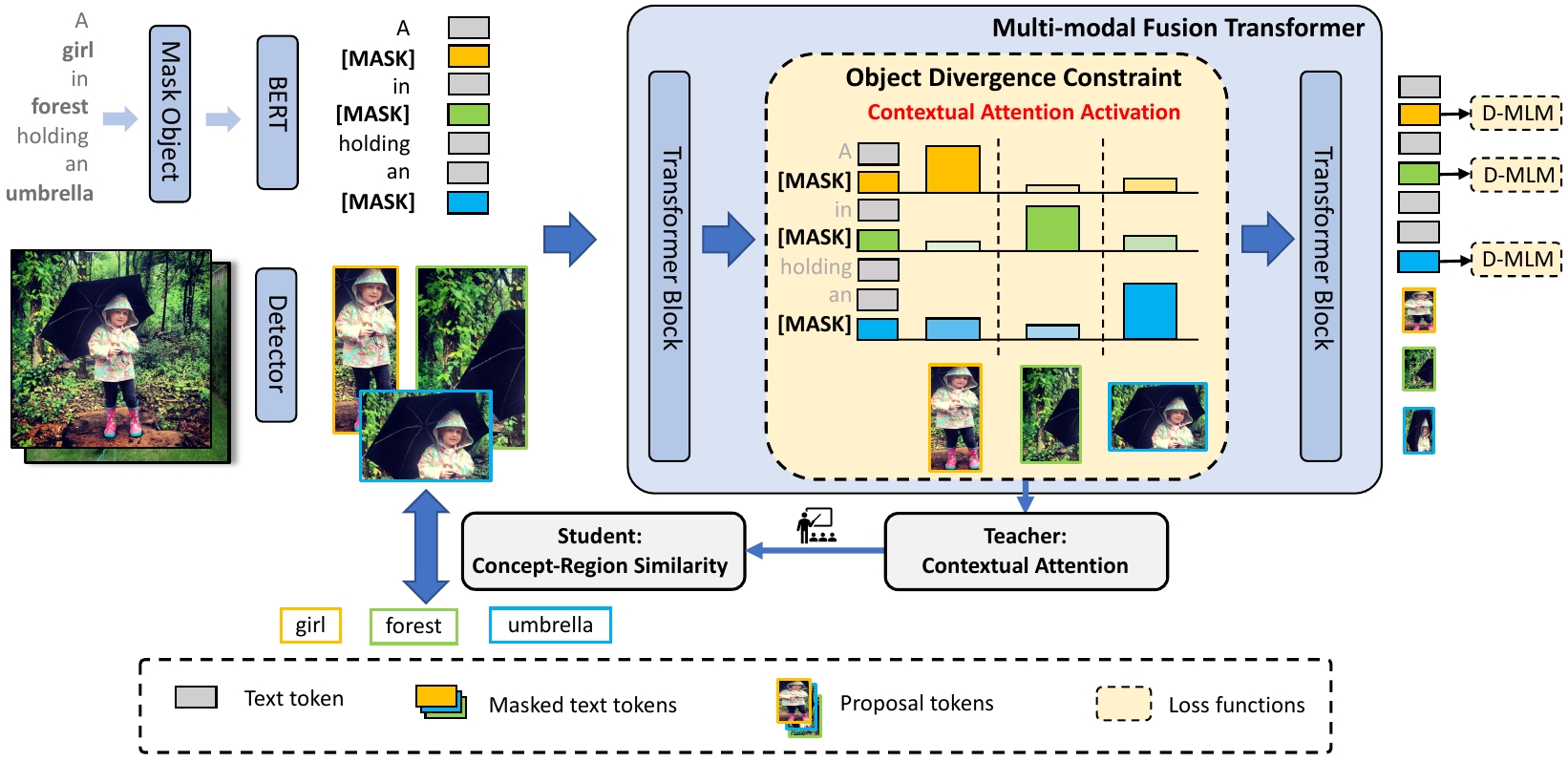}
  \caption{
  \textbf{Method overview}. We only present the training process for image-caption data. The training process for detection data is similar to that of traditional object detection.
  We distill the contextual knowledge learned in the teacher fusion transformer with our diverse multi-modal masked language modeling (D-MLM) to the student detector. The object divergence constraint encourages diverse attention activation for each concept in the fusion transformer, enabling the fusion transformer to focus on region-level visual contexts. We use the diverse contextual attention in the fusion transformer to guide the contrastive concept-region similarity matching in the detector. In practice, we copy a caption several times and only mask one concept in each replication.
  }
    \label{fig:method}
\end{figure*}

\textbf{Multi-modal Masked Language Modeling (MLM).}
In the multi-modal alignment task, MLM takes both image region features $R^I$ and the corresponding masked caption word tokens $W^T$ as inputs. As an example, $\{r_{1}, \dots, r_{n_{I}}, w_{1}, \dots, \mathrm{[MASK]}, \dots, w_{n_{T}}\}$. 
During training, each caption word is masked by a $\mathrm{[MASK]}$ token with a certain probability ~\cite{vilbert, bert}. Each $\mathrm{[MASK]}$ token is used to predict the corresponding masked word $w_{k}$ by
\begin{equation}
\mathcal{L}_\mathrm{mlm}=\mathcal{L}_\mathrm{CE}(F(\mathrm{[MASK]}), w_{k}),
\label{eqn:loss_mlm}
\end{equation}
where $F$ denotes a multi-modal fusion transformer. 

\subsection{Multi-modal Contextual Knowledge Distillation}
In this subsection, we briefly outline the proposed multi-modal contextual knowledge distillation framework. The detailed introduction on each sub-method can be found in Sec.\,\ref{sec:teacher}, Sec.\,\ref{sec:student}, and Sec.\,\ref{sec:overall_loss}.
As shown in Fig.\,\ref{fig:method}, our approach is based on a vision-language fusion framework. The multi-modal fusion transformer is treated as a teacher model, while the preceding detector is treated as a student model. Our multi-modal contextual knowledge distillation is mainly comprised of the learning strategies of the two models. 

For the teacher model, we conduct diverse multi-modal masked language modeling (D-MLM) to learn the contextual knowledge. Concretely, given an image and its corresponding caption, we first extract the region proposal tokens and partially masked text tokens using the student detector and a pretrained frozen text encoder~\cite{bert, clip}, respectively. The number of region proposals is set to $128$ per image. Both the region proposals and partially-masked text tokens are then fed into a multi-modal fusion transformer supervised by D-MLM to learn multi-modal contextual relationship. The D-MLM is introduced in detail in Sec.\,\ref{sec:teacher}.

For the student detector, we supervise its concept-region similarity with the context-aware attention scores learned in the teacher model. Specifically, we first extract the attention scores of masked concepts to each region proposal token in the last self-attention layer of the fusion transformer. Then, the extracted attention scores are used to supervise the concept-region similarity via our modified attention-based contrastive loss in Eqn.\,(\ref{eqn:new_ground_score}). After training, our approach drops all additional modules except the student detector, thus bringing no extra parameters and computational cost during inference.

We elaborate on the training details of the teacher and student models in Sec.\,\ref{sec:teacher} and Sec.\,\ref{sec:student}. The overall training schedule is summarised in Sec.\,\ref{sec:overall_loss}.

\subsection{Teacher Model: Multi-Modal Contextual Knowledge Learning}
\label{sec:teacher}
\textbf{Input token preparation.}
Given an image $I$ and its corresponding caption $T$ with words $W^{T}=\{w_{i}\}_{i=1}^{|W^{T}|}$, we respectively prepare the text and vision inputs of the fusion transformer. 

 For text inputs, we recognize the object concepts $C^{T}=\{c_{i}\}_{i=1}^{|C^{T}|} \subseteq W^{T}$ by a language parser~\cite{parser}. For example, $T = $ ``A girl under an umbrella'', $W^{T} = \{\textrm{``A''}, \textrm{``girl''}, \textrm{``under''}, \textrm{``an''}, \textrm{``umbrella''}\}$, and $C^{T} = \{\textrm{``girl''},\textrm{``umbrella''}\}$. 
 Instead of randomly masking in vanilla MLM, here we only mask one object concept in $C^{T}$ with a 100\% probability (\emph{e.g.}, ``A girl under an [MASK]''). 
To acquire all masked results of a caption with multiple concepts, we copy the caption for multiple times and only mask one concept at each time.
 We use the encoded tokens $Q$ of the partially masked caption as the text inputs, namely,
\begin{equation}
    Q = ( W^{T} \setminus C^{T} ) \cup \{M^{(c_{i})} | c_{i} \in C^{T} \},
    \label{eqn:text_input}
\end{equation}
where $M^{(c_{i})}$ denotes the masked token of concept $c_{i}$.

For vision inputs, we first use the student detector to extract the features of region proposals $R^{I}=\{r_i\}_{i=1}^{n_{I}}$ according to the image $I$. Then we coarsely pre-filter the noise proposals based on concept-region similarity, namely, 
\begin{equation}
    P^{(c_{i})}=\{r_{j}|\langle r_{j}, c_{i}\rangle \geq S_{\delta_{K}}^{(c_{i})}, r_{j} \in R^{I} \}, 
    \label{eqn:candidate_prop}
\end{equation}
where $P^{(c_{i})}$ denotes the preserved region proposal tokens of the concept $c_{i}$. $S_{\delta_{K}}^{(c_{i})}$ is the $K$-th highest value in $\langle r_{*}, c_{i}\rangle$. This function serves to select $K$ region proposals for each concept. 
Hyper-parameter $|P^{(c_{i})}|=K$ is set to a relatively large value to cover the optimal region proposals. Then the preserved regions of all concepts are concatenated as the vision inputs, namely:
\begin{equation}
    P = P^{(c_{1})} \cup P^{(c_{2})} \cup \dots \cup P^{(c_{|C^{T}|})}.
    \label{eqn:vision_input}
\end{equation}

Finally, we use the partially masked text tokens $Q$ and pre-filtered region proposal tokens $P$ as the joint inputs $\{Q,P\}$ of the fusion transformer.

\textbf{Diverse multi-modal masked language modeling.}
As described in Sec.\,\ref{sec:intro} and Fig.\,\ref{fig:vanilla_mlm}, vanilla MLM is not sensitive to region-level information.
%
%
This problem mainly results from the intrinsic limitation of vanilla MLM that simply predicts the masked words based on coarse contextual knowledge, while does not necessarily pursue the distinct and local information for individual concepts.
As a result, the fusion transformer tends to learn visual context from the similar global region proposals for all masked words.
To this end, we propose a diverse multi-modal masked language modeling (D-MLM) strategy to learn fine-grained region-level contextual knowledge. Specifically, we propose an \textbf{object divergence constraint}, as an auxiliary constraint upon the vanilla MLM for detection frameworks, to encourage different masked concepts to focus on essential local regions. 
The object divergence constrain regularizes the multi-modal fusion transformer to activate diverse attention for different masked concepts. 

\begin{figure}[t]
    \centering
    \includegraphics[width=0.45\textwidth]{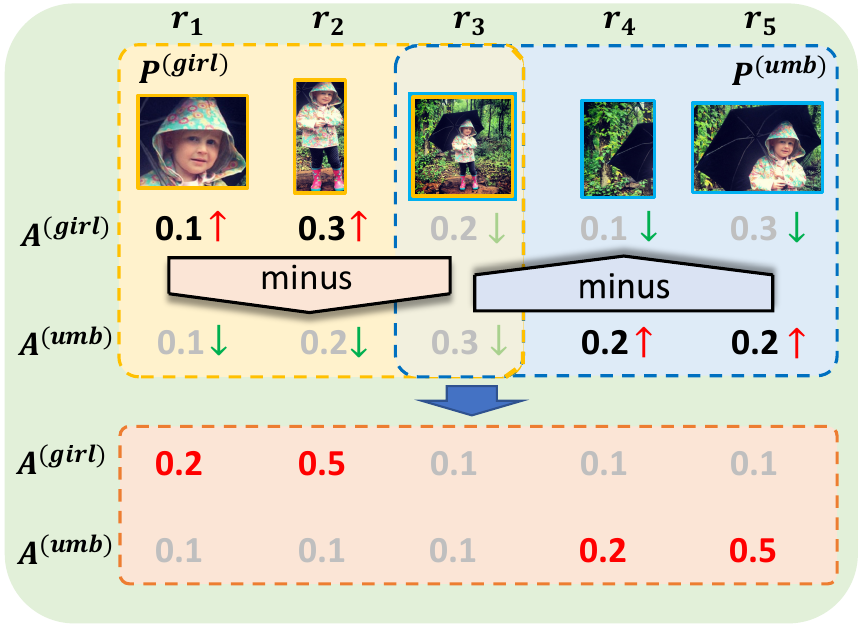}
    \caption{
    Illustration of the object divergence loss in Eqn.\,(\ref{eqn:loss_div}).
    Take the concept ``girl'' as example, with its corresponding pre-filtered region proposals $P^{(girl)}$ in Eqn.\,(\ref{eqn:candidate_prop}), \emph{i.e.}, the yellow box, we compute the attention difference $A^{(girl)}-A^{(umb)}$ within $P^{(girl)}$.
    By maximizing the difference, $A^{(girl)}$ scores on the exclusive proposals $r_{1}$ and $r_{2}$ of ``girl'' are enlarged, while $A^{(umb)}$ scores on $r_{1}$ and $r_{2}$ are curbed. Similar constraint is conducted on ``umbrella''. The scores on the shared proposal $r_{3}$ are curbed due to attention normalization (the sum of all scores always equals to one).
    The attention results after training with our object divergence loss is shown in the bottom orange box.
    }
    \label{fig:loss_div}
\end{figure}

Fig.\,\ref{fig:loss_div} concisely illustrates the computation procedure of object divergence constraint.
Given the vision and text inputs defined in Eqns.\,(\ref{eqn:text_input}) and (\ref{eqn:vision_input}), 
we first get the attention scores of masked concepts to each region proposal token in the last self-attention layer of the fusion transformer:
\begin{equation}
A_{j}^{(c_{i})}=\langle W_{q} M^{(c_{i})}, W_{k} r_{j} \rangle, \quad j=1, 2, \dots, |P|,
\label{eqn:attn}
\end{equation}
where $W_{q}$ and $W_{k}$ denote the linear projections of the query and key in the attention module. 
Then, the divergence loss with threshold $\alpha=0.5$ is defined as
\begin{equation}
\mathcal{L}_\mathrm{div}=[\alpha-\frac{1}{|C^{T}|} \sum_{c_{i}, c_{j} \in C^{T}} \sum_{r_{k}\in P^{(c_{i})}} (A_{k}^{(c_{i})} - A_{k}^{(c_{j})})]_{+},
\label{eqn:loss_div}
\end{equation}
where $[*]_{+}=max\{*, 0\}$.

As illustrated in Fig.\,\ref{fig:loss_div}, the object divergence loss in Eqn.\,(\ref{eqn:loss_div}) encourages each concept to activate sparse attention on its exclusive proposals, namely, the exclusive proposals always get high attention activation. For example, the attention score on the proposal $r_{2}$ of the concept ``girl'' and  the score on the proposal $r_{5}$ of ``umbrella'' are increased, respectively.
For those global proposals containing multiple instances of different concepts, 
the divergence constraint does not take direct effect on them, \emph{e.g.}, $(A_{3}^{(girl)}-A_{3}^{(umb)})+(A_{3}^{(umb)}-A_{3}^{(girl)})=0$ in Fig.\,\ref{fig:loss_div}, according to Eqn.\,(\ref{eqn:loss_div}). However, we empirically observe that 
the attention scores of the shared proposals are curbed by the increasing scores of the exclusive proposals, since the sum of the attention scores always equals to one, namely, attention normalization.
To this end, each concept learns diverse attention activation and focuses on its own region proposals (see Fig.\,\ref{fig:vis_select}).

Together with the multi-modal masked language modeling loss modified from Eqn.\,(\ref{eqn:loss_mlm}), we get our diverse multi-modal masked language modeling loss as:
\begin{equation}
\mathcal{L}_\mathrm{divmlm}= \mathcal{L}_\mathrm{div} + \frac{1}{|C^{T}|} \sum_{c_{i} \in C^{T}} \mathcal{L}_\mathrm{CE}(F(M^{(c_{i})}), c_{i}),
\label{eqn:loss_dymlm}
\end{equation}
where the second term is the MLM loss on all concepts. The teacher model with multi-modal fusion transformer is trained via the diverse multi-modal masked language modeling in Eqn.\,(\ref{eqn:loss_dymlm}).

\subsection{Student Model: Context-Aware Concept-Region Matching}
\label{sec:student}
We build the student detector based on a baseline open-vocabulary object detector trained via $\mathcal{L}_\mathrm{ovd}$ in Eqn.\,(\ref{eqn:loss_baseline}).
In addition to traditional open-vocabulary detection training, we introduce our distillation training strategies to transfer the contextual knowledge in the teacher model to the student detector, as described in the following part.

\textbf{Contextual knowledge distillation.}
We reuse the attention scores $A_{j}^{(c_{i})}$ extracted via Eqn.\,(\ref{eqn:attn}) to conduct our contextual knowledge distillation. 
The learned attention scores are acquired by jointly modeling the contexts in both modalities, thereby implicitly containing multi-modal contextual knowledge. 
The context-aware attention scores are used to supervise the concept-region similarity. Specifically, we replace the similarity weights defined in Eqn.\,(\ref{eqn:sim_weight}) with corresponding attention scores, and form new local grounding scores over Eqn.\,(\ref{eqn:ground_score}), as:
\begin{equation}
\langle C^{T},R^{I}\rangle_A=\frac{1}{|C^{T}|} \sum_{c_i \in C^{T}} \sum_{r_j \in P} A_{j}^{(c_{i})} \left\langle c_i, r_j\right\rangle. 
\label{eqn:new_ground_score}
\end{equation}
Notably, the attention $A_{j}^{(c_{i})}$ is only available for positive image-caption pairs, \emph{i.e.}, images and their corresponding captions. To this end, we modify the loss in Eqn.\,(\ref{eqn:loss_contrast}) based on our new local grounding scores in Eqn.\,(\ref{eqn:new_ground_score}), and form our new attention-based contrastive loss as
\begin{equation}
\mathcal{L}_\textrm{con}^{A}(I, T)=-\log \frac{\exp \langle C^{T},R^{I}\rangle_A}{\sum_{T^{\prime} \in \mathcal{B}_T }{ \exp \left\langle C^{T^{\prime}}, R^{I} \right\rangle_S}}.
\label{eqn:loss_contr_attn}
\end{equation}
where $\langle C^{T},R^{I}\rangle_A$ and $\langle C^{T},R^{I}\rangle_S$ are computed via Eqn.\,(\ref{eqn:new_ground_score}) and Eqn.\,(\ref{eqn:ground_score}), respectively.
Here we only use the concept words instead of all caption words to compute the loss, in order to directly supervise the concept-region similarity. 
During training, the loss in Eqn.\,(\ref{eqn:loss_contr_attn}) serves to align the concept-region similarity $\left\langle c_i, r_j\right\rangle$ to the corresponding attention score $A_{j}^{(c_{i})}$, thus transferring the contextual knowledge from the fusion transformer to the preceding detector.

To this end, the final distillation loss is defined as:
\begin{equation}
   \mathcal{L}_\mathrm{distill}=\mathcal{L}_\textrm{con}^{A}(I, T)+\mathcal{L}_\textrm{con}^{A}(T, I),
    \label{eqn:loss_align}
\end{equation}
which corresponds to Eqn.\,(\ref{eqn:loss_cap}).

\subsection{Overall Training Schedule}
\label{sec:overall_loss}
We train MMC-Det via two stages. The first stage is pretraining stage, which serves to acquire good initialization of the teacher model before multi-modal contextual knowledge distillation. This stage is conducted via
\begin{equation}
\mathcal{L}_\textrm{stage1}=\mathcal{L}_\textrm{ovd}+\mathcal{L}_\mathrm{divmlm}.
\label{eqn:stage1}
\end{equation}
The second stage is the distillation stage with all losses, namely,
\begin{equation}
\mathcal{L}_\textrm{stage2}=\mathcal{L}_\textrm{ovd}+\mathcal{L}_\mathrm{divmlm}+\mathcal{L}_\mathrm{distill}.
\label{eqn:overall_loss}
\end{equation}
It is worth noting that the weight factors are omitted for concision.



\section{Experiments}
\label{sec:exp}
\subsection{Setup}
\label{sec:setup}
\textbf{Datasets and Metric.}
We evaluate our method on the challenging and widely-used COCO Object~\cite{coco} and LVIS v1~\cite{lvis} datasets. 
Following~\cite{ovrcnn}, we use modified COCO 2017 training and validation splits for training and evaluation, respectively. The training set is represented as COCO Base, which includes $107,761$ images with $48$ base classes $Y_\textrm{base}$, while the validation set contains $4,836$ images with the whole test categories $Y_{\textrm{test}}$ including both $Y_{\textrm{base}}$ and extra $17$ novel classes $Y_{\textrm{novel}}$. The LVIS contains totally 1,203 categories. Following~\cite{detic}, we use the 337 rare classes as novel categories, while recognizing the remaining 866 common and frequent classes as base categories. We use COCO Captions dataset~\cite{coco-caption} and Conceptual Captions~\cite{cc3m} (CC3M) as the image-caption data when evaluating on COCO and LVIS datasets, respectively. In addition, we use Objects365~\cite{objects365} and OpenImages~\cite{openimages} datasets to conduct cross-dataset evaluation, in order to verify the generalization of our models. All datasets used in this work are commonly used evaluation benchmarks in previous OVD works.
We parse each caption in the training caption set by Scene-Graph-Parser~\cite{parser,parser_link}.
We adopt standard object detection metric AP$50$ (average precision at an intersection over union of $0.5$) for COCO evaluation and Average Precision (AP) on segmentation for LVIS evaluation. The AP on rare, common, and frequent classes are denoted as $\textrm{AP}_{\textrm{r}}$, $\textrm{AP}_{\textrm{c}}$, and $\textrm{AP}_{\textrm{f}}$, respectively. 
Since the base categories are supervised by box annotations, we mainly analyze the performance on novel classes for open-vocabulary ability.

\textbf{Implementation Details.}
We use Faster R-CNN~\cite{faster_rcnn} with RN50-C4 as our default detector configuration. For evaluation on LVIS benchmark, we follow Detic~\cite{detic} to use CenterNetv2~\cite{centernet} with RN50-FPN as our backbone.
We use a pretrained and frozen BERT-Base-Uncased~\cite{bert} as our language encoder.
Notably, we use the BERT embeddings (\emph{i.e.}, the hidden states before the BERT pooling layer) as the outputs of the text encoder.
The classifier weights are also replaced with the fixed BERT embeddings of corresponding category texts for open-vocabulary classification.
For the multi-modal fusion transformer, a $6$-layer and $8$-head transformer encoder~\cite{transformer} is adopted.
We adopt CLIP text encoder~\cite{clip} for LVIS training and observe better scalability. To enable masked language modeling, we remove the last pooling layer, and utilize ``\#'' as the mask token. 

To train MMC-Det, 
we set the number of preserved proposals of each concept $|P^{(c_{i})}|=10$ in Eqn.~(\ref{eqn:candidate_prop}) during pre-filtering.
The object divergence constraint is applied to the last layer of the multi-modal fusion transformer to encourage low-level feature encoding in the shallow layers. We use the averaged attention scores among all heads in the last self-attention layer of the multi-modal fusion transformer to conduct the contextual knowledge distillation.


\textbf{Training schedules.}
We train MMC-Det via two stages as described in Sec.\,\ref{sec:overall_loss}. The pretraining stage  is trained for $\sim$$1\times$ schedule. 
The distillation stage is trained for another $\sim$$1\times$ schedule.

\textbf{Computation cost}. We train all the models on 8 V100 GPUs. The training of COCO includes 17 hours for the pretraining stage and 10 hours for the distillation stage. The training of LVIS includes 25 hours for the pretraining stage and another 25 hours for the distillation stage.

\textbf{Weight factors of losses}. The weight factors of the losses serve to regularize the losses to relatively similar scales. For the losses of Eqn.\,(\ref{eqn:stage1}) in the pretraining stage, we set the weights of $\mathcal{L}_\textrm{det}$, $\mathcal{L}_\textrm{cap}$, $\mathcal{L}_\mathrm{img}$, and $\mathcal{L}_\mathrm{divmlm}$ to $1.0$, $0.1$, $0.1$, and $0.1$.
For the losses of Eqn.\,(\ref{eqn:overall_loss}) in the distillation stage, we set the weights of $\mathcal{L}_\textrm{det}$, $\mathcal{L}_\textrm{cap}$, $\mathcal{L}_\mathrm{img}$, $\mathcal{L}_\mathrm{divmlm}$, and $\mathcal{L}_\mathrm{distill}$ to $1.0$, $0.1$, $0.1$, $0.1$, and $0.1$.
More reasonable weight settings may further benefit the performance.

\subsection{Comparison to Open-Vocabulary Detectors}

\begin{table}[t]
\centering
\caption{OVD results on COCO. 
All models are trained with the same image-text dataset COCO Captions~\cite{coco-caption}.
}
\label{tab:main_result}
\resizebox{0.93\linewidth}{!}{
\setlength{\tabcolsep}{0.6mm}{
\begin{tabular}{l|cc|ac}
\toprule
               & \multicolumn{2}{c|}{Detector training} & \multicolumn{2}{c}{COCO AP$50$ (\%)} \\
Method         & Backbone  & Box generator & Novel        & \g{All} \\
\midrule
Supervised (Base) & RN50-C4   & COCO Base                 & 0.3      & \g{39.2}      \\
\midrule
                \multicolumn{5}{c}{Weakly-supervised object detection}              \\
\midrule
WSDDN~\cite{wsddn}                   & -   & -                 & 20.5       & \g{24.6}       \\
Cap2Det~\cite{detic}                   & -   & -                 & 20.3        & \g{20.1}       \\
\midrule
                \multicolumn{5}{c}{Open-vocabulary object detection}              \\
\midrule
OVR-CNN~\cite{ovrcnn}                & RN50-C4   & COCO Base                 & 22.8    & \g{39.9}       \\
HierKD~\cite{hierkd}                  & RN50-C4   & COCO Base                 & 20.3     & \g{43.2}       \\
Detic\tablefootnote{Reproduced.\label{repr}}~\cite{detic}                   & RN50-C4   & COCO Base                 & 27.8     & \g{45.0}       \\
PB-OVD~\cite{pb-ovd}                 & RN50-C4   & COCO Base                 & 30.8      & \g{42.1}       \\
PromptDet~\cite{promptdet}         & RN50-FPN   &  COCO Base         & 26.6    & \g{50.6}       \\
ViLD~\cite{vild}                     & RN50-FPN  & COCO Base                 & 27.6    & \g{51.3}       \\
\midrule
\textbf{MMC-Det (ours)}                               & RN50-C4   & COCO Base                 & \textbf{33.5}   &   \g{47.5}       \\
\midrule\midrule
RegionCLIP~\cite{regionclip}         & RN50-C4   & LVIS~\cite{lvis}                      & 26.8    & \g{47.5}       \\
VL-PLM~\cite{vlplm}                  & RN50-FPN  &  LVIS~\cite{lvis}          & 34.4        & \g{53.5}      \\
\bottomrule
\end{tabular}
}
}
\end{table}

\textbf{Open-vocabulary COCO.}
As illustrated in Tab.~\ref{tab:main_result}, we compare MMC-Det with both advanced weakly-supervised and open-vocabulary detection methods.
Compared under the same backbone with RN50-C4 and box generator pretrained on COCO Base, our context-based MMC-Det outperforms the previous methods without multi-modal contextual knowledge by an obvious margin, \emph{e.g.}, $+5.7\%$ over Detic~\cite{detic}.
Meanwhile, MMC-Det shows comparable performance when compared with OVD methods~\cite{regionclip, vlplm} with box generators trained upon the stronger detection dataset LVIS. It is worth noting that there may exist information leak in RegionCLIP~\cite{regionclip} and VL-PLM~\cite{vlplm} on their COCO evaluation. The reason is that their box generators are pretrained on the LVIS dataset, which shares the same images with COCO and contains novel categories. As shown in Tab.~\ref{tab:LVIS}, when the box generator is pretrained under LVIS, MMC-Det outperforms VL-PLM by a large margin.
We notice that stronger backbones (\emph{e.g.}, RN50-FPN) can result in better overall performance, which means potential performance gains of MMC-Det. For fair comparison, we choose RN50-C4 as baseline backbone. 


\begin{table}[t]
   \centering
    \caption{OVD results on LVIS dataset~\cite{lvis}. All models are trained with the same image-text dataset CC3M~\cite{cc3m}.}
    \label{tab:LVIS}
    \resizebox{\linewidth}{!}{
    \begin{tabular}{l|cccccc}
        \toprule
        Method & Backbone  & $\text{AP}_{\text{r}}$ & \color{gray}{$\text{AP}_{\text{c}}$} &  \color{gray}{$\text{AP}_{\text{f}}$} &\color{gray}{$\text{AP}$} \\
        \midrule
        Supervised (Base)  &  RN50-FPN   & 16.3 &  \g{31.0}  &  \g{35.4} &\color{gray}{30.0}  \\
        \midrule
        WSDDN~\cite{wsddn} &  -    & 16.5  &   \g{-}   &  \g{-}  &\color{gray}{30.0}  \\
        \midrule
        ViLD~\cite{vild}   & RN50-FPN   & 16.6  &  \g{24.6}  &  \g{30.3} &\color{gray}{25.5}  \\
        RegionCLIP~\cite{regionclip} & RN50-C4   & 17.1  & \g{27.4}   & \g{34.0}  &\color{gray}{28.2}  \\
        VL-PLM~\cite{vlplm} & RN50-FPN   & 17.2 &  \g{23.7}  &  \g{35.1} & \color{gray}{27.0} \\
        Detic\textsuperscript{\ref {repr}}~\cite{detic} & RN50-FPN    & 19.3  &  \g{30.8}   & \g{34.9}  &\color{gray}{30.4}  \\
        DetPro~\cite{detpro} &  RN50-FPN   & 19.8 &  \g{25.6}  &  \g{28.9} &\color{gray}{25.9} \\
        \midrule
         \rowcolor{Tabcolor} \textbf{MMC-Det (Ours)} & RN50-FPN     & \textbf{21.1}  & \color{gray}{\textbf{30.9}}  & \color{gray}{\textbf{35.5}} &\color{gray}{\textbf{31.0}} \\
        \bottomrule
    \end{tabular}
    }
\end{table}

\textbf{Improvement gap between COCO and LVIS.} We observe a narrowing improvement gap between the two evaluated datasets. For example, our improvement on LVIS over Detic~\cite{detic} is +1.8\% in Tab.~\ref{tab:LVIS} while  +5.7\% on COCO in Tab.~\ref{tab:main_result}. Since all novel categories in LVIS are actually rare classes, the amount of instances of the rare categories is extraordinarily limited even in large image-caption datasets. The challenging long tail nature may impede the discriminative learning. As shown in Tab.~\ref{tab:LVIS}, many OVD methods only gain sight improvement (\emph{e.g.}, +0.3\% and +0.8\% of ViLD and RegionCLIP over the base-class-supervised model). Thus, the improvement on LVIS is enough to verify the effectiveness of our context-based approach. Meanwhile, our approach scales well on larger categories (\emph{e.g.}, 1,203 classes in LVIS) and data sizes (\emph{e.g.}, 3 million image-text pairs in CC3M).

\begin{table}[t]
   \centering
    \caption{Cross-dataset evaluation. All methods are trained on LVIS~\cite{lvis} and directly evaluated on other datasets.}
    \label{tab:cross}
    \resizebox{\linewidth}{!}{
    \begin{tabular}{l|ccc}
        \toprule
    Method                  & COCO & Objects365 & OpenImages \\
        \midrule
        ViLD-text~\cite{vild}   & 43.4 &  15.8  &     -          \\ 
        Detic-base~\cite{detic} & 55.3 &  19.2  &     36.1       \\
        \midrule
        DetPro~\cite{detpro}    & 53.8 &  18.8  &     -          \\
        ViLD~\cite{vild}        & 55.6 &  18.2  &     -          \\
        Detic~\cite{detic}      & 55.9 &  20.4   &    37.2        \\
        \midrule
        \textbf{MMC-Det (Ours)}           & \textbf{56.4} &  \textbf{21.4}  &  \textbf{38.6} \\
        \bottomrule
    \end{tabular}
    }
\end{table}

\textbf{Cross-dataset evaluation.}
We provide the cross-dataset evaluation results of MMC-Det in Tab.~\ref{tab:cross}. We utilize the MMC-Det model trained on LVIS in Tab.\,\ref{tab:LVIS} and conduct zero-shot evaluation on other datasets. We report AP50 on the COCO~\cite{coco}, Objects365~\cite{objects365}, and OpenImages~\cite{openimages} datasets. ViLD-text~\cite{vild} and Detic-base~\cite{detic} are the baselines supervised by base categories for ViLD and Detic respectively. Our approach is built on the CenterNetv2~\cite{centernet} backbone of Detic and shows favourable generalization when directly transferred to new datasets without any dataset-speciﬁc ﬁnetuning.

\subsection{Multi-modal Context Learning in OVD}

\begin{table}[t]
\centering
\caption{Effect of caption types in MMC-Det. 
Models are trained followed by the training schedule in Sec.\,\ref{sec:overall_loss}. The results on COCO are reported.
}
\label{tab:caption}
\resizebox{0.95\linewidth}{!}{
\setlength{\tabcolsep}{1.5mm}{
\begin{tabular}{c|acc|acc}
\toprule
\multirow{2}{*}{Caption type}                & \multicolumn{3}{c|}{Pretraining} &  \multicolumn{3}{c}{Distillation}\\
                    & Novel & \g{Base} & \g{All} & Novel & \g{Base} & \g{All}  \\
\midrule
Only concepts       &  28.8 &  \g{\textbf{52.4}}  & \g{\textbf{46.2}} & 30.5  & \g{52.2}  & \g{46.5}   \\
Single word         &  28.7 &  \g{52.0} & \g{46.0}& 26.8 &\g{49.3}& \g{43.4}  \\
Full caption        &  \textbf{31.0}   &  \g{51.4} & \g{46.1} &  \textbf{33.5} & \g{\textbf{52.4}} & \g{\textbf{47.5}} \\
\bottomrule
\end{tabular}
}
}
\end{table}

\textbf{Contextual knowledge in captions.}
We explore how the contextual knowledge in captions promotes open-vocabulary detection.
The contextual knowledge in captions is implicitly embraced by the relation words like ``under'' and ``by'', and interaction among concepts.
Thus, we construct three types of captions to train MMC-Det in Tab.~\ref{tab:caption}: (1) only concepts: concepts combined with ``,'' (\emph{e.g.}, ``girl, umbrella, ...''); (2) single word: only one concept word (\emph{e.g.}, ``umbrella''); (3) full caption (\emph{e.g.}, ``a girl under an umbrella'').
The performance after distillation of combining concepts without relation words (``only concepts'') drops by an obvious margin compared to full captions, \emph{i.e.}, -3.0$\%$ AP$50$ on novel classes.
We observe that utilizing a single concept word (``single word '') as the caption further degrades the performance and even impairs the pretrained features after distillation, \emph{i.e.}, -1.9\% AP$50$ when comparing results in the distillation and pretraining stages. 
The results quantitatively indicate that OVD indeed benefits from the reasonable usage of  contextual knowledge in relation words and concept interaction.

\begin{figure*}[t]
    \centering
    \includegraphics[width=\textwidth]{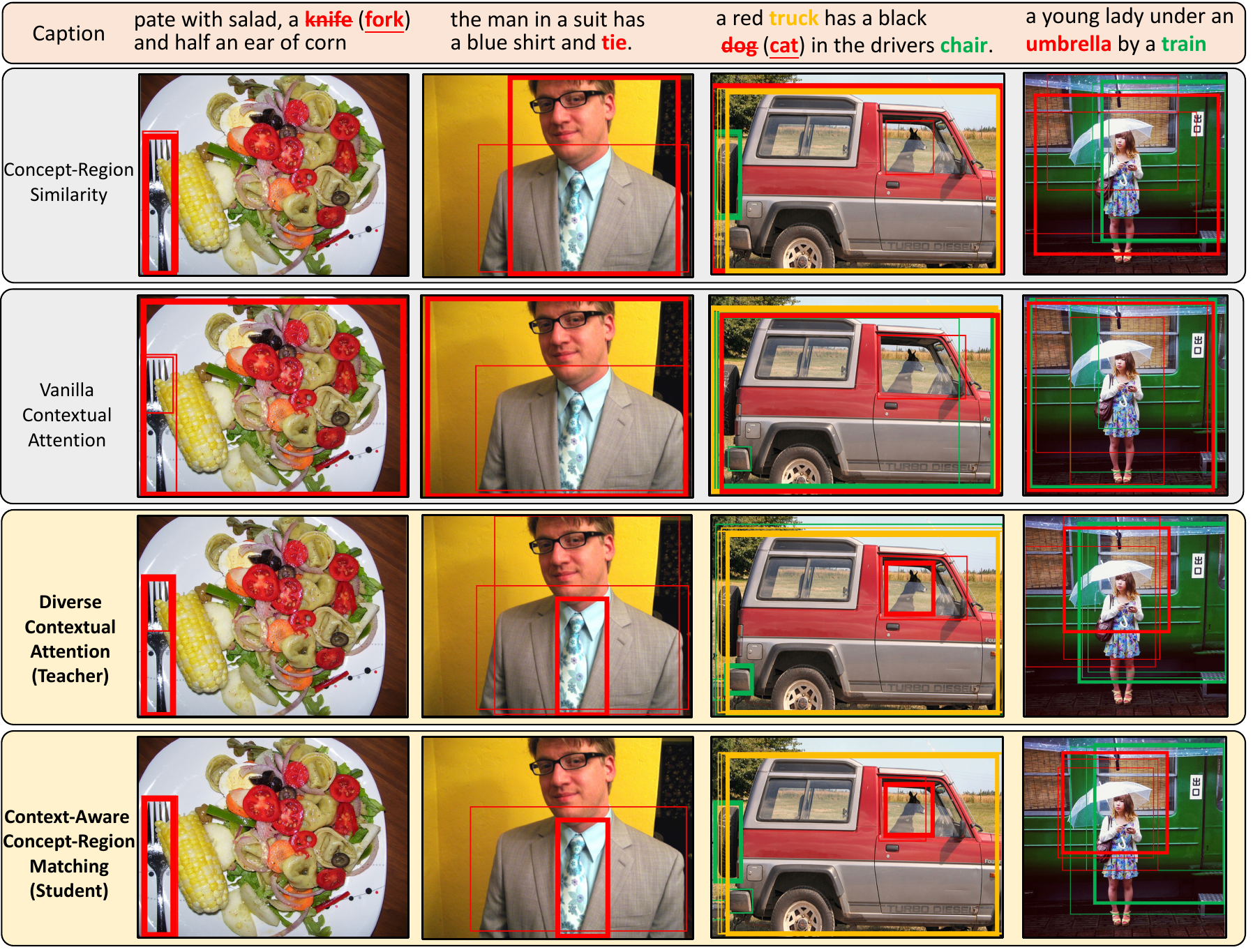}
    \caption{
    Visualization of the concept-region matching results of different approaches. Each concept is assigned to at most three proposals via different strategies, namely, simple concept-region dot product similarity without the contextual knowledge distillation, vanilla contextual attention without our object divergence constraint, our diverse contextual attention in the teacher fusion transformer, and the concept-region matching of our student detector. For clarity, we bold the top-1 proposal of each concept. The masked concepts are colored. All novel concepts are in red color. If a concept is mispredicted by the D-MLM, we mark it as \textbf{\sout{groundtruth} (\uline{prediction})}.
    }
    \label{fig:vis_select}
\end{figure*}

\textbf{Qualitative visualization.}
To better understand the contextual knowledge learned in our multi-modal contextual knowledge distillation framework, we visualize the concept-region matching results of different approaches in Fig.~\ref{fig:vis_select}. The first row shows the result of traditional concept-region similarity, which directly conduct concept-region matching via simple similarity dot product. 
Instead, we use the attention scores in the fusion transformer to select top-3 proposals for each concept. We show the results of the vanilla contextual attention scores in a fusion transformer trained with vanilla MLM in the second row. It is illustrated that vanilla contextual attention always highly activates on similar global regions to coarsely integrate the contextual knowledge, indicating the insensitivity to region-level information of vanilla contextual attention. 
To this end, we evaluate our approach in the third and fourth rows. The third row shows the result using our diverse contextual attention in the teacher fusion transformer trained with D-MLM, while the fourth row presents the result of the student detector after our distillation training.
Compared to concept-region similarity in the previous OVD methods (the first row), our diverse contextual attention activation with object divergence constraint (the third row) has more discriminative perception on novel categories (see ``tie'' and ``umbrella''), thus proving the effectiveness of the contextual knowledge modeling ability of the fusion transformer with our diverse multi-modal masked language modeling (D-MLM) strategy.
We also observe that the detector learned via our contextual knowledge distillation strategy (the forth row) can focus on more accurate region proposals for each concept, therefore proving the effectiveness of our contextual knowledge distillation strategy.

\begin{table}[t]
\centering
\caption{Effect of noise removal in MMC-Det.}
\label{tab:noise_removal}
\setlength{\tabcolsep}{1.5mm}{
\begin{tabular}{c|ac|ac}
\toprule
\multirow{2}{*}{Method} & \multicolumn{2}{c|}{COCO}  & \multicolumn{2}{c}{LVIS}  \\
 & Novel & \g{All}  & Rare   &   \g{All}    \\
\midrule
w/ Noise Removal & 33.5    &  \g{47.5}   & 21.1  & \g{31.0} \\
w/o Noise Removal  & 32.5   & \g{47.0}    & 20.7 & \g{30.8} \\
\bottomrule
\end{tabular}
}
\end{table}

\textbf{Noise concept removal.}
As shown in Fig.~\ref{fig:vis_select}, the noise concepts can be recognized via simple D-MLM predictions. Take the concept ``knife'' in the first caption as an example. During D-MLM training, we mask the concept ``knife'' and feed the partially masked caption and corresponding image into the fusion transformer to conduct masked word prediction. The prediction is based on the joint contexts of images and texts. Therefore, if a concept does not actually exist in the image (\emph{e.g.}, there exists a ``fork'' instead of a ``knife'' in the image), the prediction will be inconsistent with the corresponding masked concept. We treat the concepts that are inconsistent with their predictions as noise concepts.
To this end, we remove the noise concepts when conducting contextual knowledge distillation, namely, removing the noise concepts from $C^{T}$ when calculating our attention-based contrastive loss in Eqn.\,(\ref{eqn:loss_contr_attn}) and Eqn.\,(\ref{eqn:loss_align}).
Meanwhile, we observe that some positive concepts are mis-predicted (\emph{e.g.}, ``dog'' in the third caption of Fig.\,\ref{fig:vis_select} is predicted as ``cat'') and recognized as noise concepts,  which may lead our model to neglect learning from some hard positive concepts. Tab.\,\ref{tab:noise_removal} shows the quantitative results. The results indicate that our noise removal strategy significantly boost the detection performance.

\subsection{Ablation Analysis}
\label{sec:ablation}

\begin{table}[t]
\centering
\caption{Ablation results for different losses in MMC-Det.}
\label{tab:ablation}
\resizebox{\linewidth}{!}{
\setlength{\tabcolsep}{1.5mm}{
\begin{tabular}{ccccc|ac|ac}
\toprule
\multicolumn{5}{c|}{Method} & \multicolumn{2}{c|}{COCO}  & \multicolumn{2}{c}{LVIS}  \\
$\mathcal{L}_{\mathrm{det}}$ & $\mathcal{L}_{\mathrm{cap}}$  & $\mathcal{L}_{\mathrm{img}}$ & $\mathcal{L}_{\mathrm{divmlm}}$ & $\mathcal{L}_{\mathrm{distill}}$ & Novel & \g{All}  & Rare   &   \g{All}    \\
\midrule
$\checkmark$        &                      &                     &                       &                               & 0.3    &  \g{39.2}   & 16.4  & \g{30.2} \\
$\checkmark$        &     $\checkmark$     &                     &                       &                               & 11.1   & \g{40.6}    & 18.6 & \g{30.3} \\
$\checkmark$        &     $\checkmark$     & $\checkmark$        &                       &                               & 28.5   & \g{45.7}    & 19.3 & \g{30.4} \\
$\checkmark$        &     $\checkmark$     & $\checkmark$        & $\checkmark$          &                               &  31.0  & \g{46.1}    & 19.3 & \g{30.4}       \\
$\checkmark$        &     $\checkmark$     & $\checkmark$        &           &      $\checkmark$              &  27.7  & \g{45.5}    & 18.8 & \g{30.3}       \\
$\checkmark$        &   $\checkmark$       & $\checkmark$        & $\checkmark$          & $\checkmark$                 & \textbf{33.5}  &  \g{\textbf{47.5}}   & \textbf{21.1}  & \g{\textbf{31.0}}     \\
\bottomrule
\end{tabular}
}
}
\end{table}

\textbf{Loss ablation.}
Tab.~\ref{tab:ablation} shows the OVD performance of MMC-Det by adding the losses in Sec.\,\ref{sec:overall_loss} sequentially.
It is worth noting that the first four rows are conducted only with the pretraining stage, while the last two rows are conducted with both the pretraining and distillation stages.
We first get a baseline OVD model without the fusion transformer by $\mathcal{L}_{\mathrm{det}}$, $\mathcal{L}_{\mathrm{cap}}$, and $\mathcal{L}_{\mathrm{img}}$ (the third row).
We observe that only replace vanilla MLM with our D-MLM (the fourth row) gains limited improvement over the baseline, despite the diverse attention activation shown in Fig.~\ref{fig:vis_select}. This is because the contextual knowledge that is mainly learned in the fusion transformer cannot transfer to the preceding detector without explicit supervision, thus indicating the importance of our multi-modal contextual knowledge distillation.
Meanwhile, directly conducting contextual knowledge distillation with vanilla MLM (the fifth row) greatly impairs the performance, indicating the insensitivity to region-level information of vanilla MLM.
Together with $\mathcal{L}_{\mathrm{divmlm}}$ and $\mathcal{L}_{\mathrm{distill}}$ (the last row), MMC-Det gains considerable improvement over the baseline (+5.0$\%$ AP$50$ on COCO and +1.8$\%$ AP on LVIS).

\begin{table}[t]
\centering
\caption{Effect of different layers to conduct object divergence constraint. Our default setting is in colored.}
\label{tab:loc_loss_div}
\begin{tabular}{cccccc|ccc}
\toprule
\multicolumn{6}{c|}{Transformer Layer}&\multicolumn{3}{c}{COCO AP50 ($\%$)} \\
1&2&3&4&5&6              & Novel     & \g{Base}    & \g{All} \\
\midrule
\checkmark&&&&&              &  31.9 & \g{52.1} &  \g{46.9} \\
&\checkmark&&&&              &  33.4 & \g{51.9} &  \g{47.3} \\
&&\checkmark&&&              & 33.4  & \g{52.1} &  \g{47.4} \\
&&&\checkmark&&              &  33.1 & \g{52.3} &  \g{47.5} \\
&&&&\checkmark&             &  33.2 & \g{52.2} &  \g{47.4} \\
\rowcolor{Tabcolor} &&&&&\checkmark             &  \textbf{33.5} & \textbf{\g{52.2}} &  \textbf{\g{47.5}}\\
\midrule
\checkmark&\checkmark&&&&     &  33.5 & \g{52.0} &  \g{47.4} \\
&&&&\checkmark&\checkmark          &  33.2 & \g{52.1} &  \g{47.3} \\
&&\checkmark&\checkmark&\checkmark&\checkmark    &  32.7 & \g{51.9} &  \g{47.0} \\
\bottomrule
\end{tabular}
\end{table}

\textbf{Location of object divergence loss.}
To explore the effect of the location of object divergence constraint, we conduct our object divergence constraint in different transformer layers, and utilize the attention scores in corresponding layers to conduct our multi-modal contextual knowledge distillation. Results are demonstrated in Tab.\,\ref{tab:loc_loss_div}. We observe that conducting object divergence constraint in the first layer causes great performance degradation, \emph{e.g.}, $-1.6\%$ AP50 on novel classes over the result of the last layer. The reason is that the features in the early layers are not sufficiently encoded. We also observe that MMC-Det does not benefit from multi-layer fusion, \emph{i.e.}, conducting the object divergence constraint in multiple layers and aggregating their attention scores for our distillation. To this end, we choose the last layer as our default setting in order to encourage low-level feature encoding in the shallow layers.

\begin{table}[t]
\centering
\caption{AP50 ($\%$) on novel classes of COCO with different thresholds of object divergence loss. The converged min and max AP50 during training are also reported. Our default setting is in colored.}
\label{tab:div_threshold}
\begin{tabular}{c|ccacc}
\toprule
Threshold $\alpha$              & 0.1     & 0.3    & 0.5       & 0.7 & 0.9 \\
\midrule
Pretraining        & 31.0 & 31.2 &   31.0  & 31.3 &    \textbf{31.5}\\
Distillation              & 33.7  & 33.4 &   33.5  & \textbf{34.0}  & 32.9 \\
\midrule
\g{Minimum}            & \g{33.0} & \g{32.9} &   \g{33.2}  & \g{33.6} &    \g{32.2}\\
\g{Maximum}            & \g{33.7} & \g{33.4} &   \g{33.8}  & \g{34.2} &    \g{33.2}\\

\bottomrule
\end{tabular}
\end{table}

\textbf{Threshold of object divergence constraint.}
The threshold $\alpha$ of object divergence loss in Eqn.\,(\ref{eqn:loss_div}) controls the extent of object divergence constraint. We analyse its effect in Tab.\,\ref{tab:div_threshold}. Results show that the model in the pretraining stage gains consistent performance improvement with larger thresholds for divergence loss. However, too large thresholds may lead to the over-confidence problem in the distillation stage. Specifically, the teacher model may highly attend to the top-$1$ proposal while neglecting auxiliary clues in other proposals, thus transferring misleading contextual knowledge to the student detector.
We also observe that our approach is relatively robust to the threshold $\alpha$ value, and performs better with reasonably large divergence, \emph{e.g.}, $34.0\%$ AP50 with the threshold of $0.7$.

\begin{figure}[t]
\centering
\includegraphics[width=0.25\textwidth]{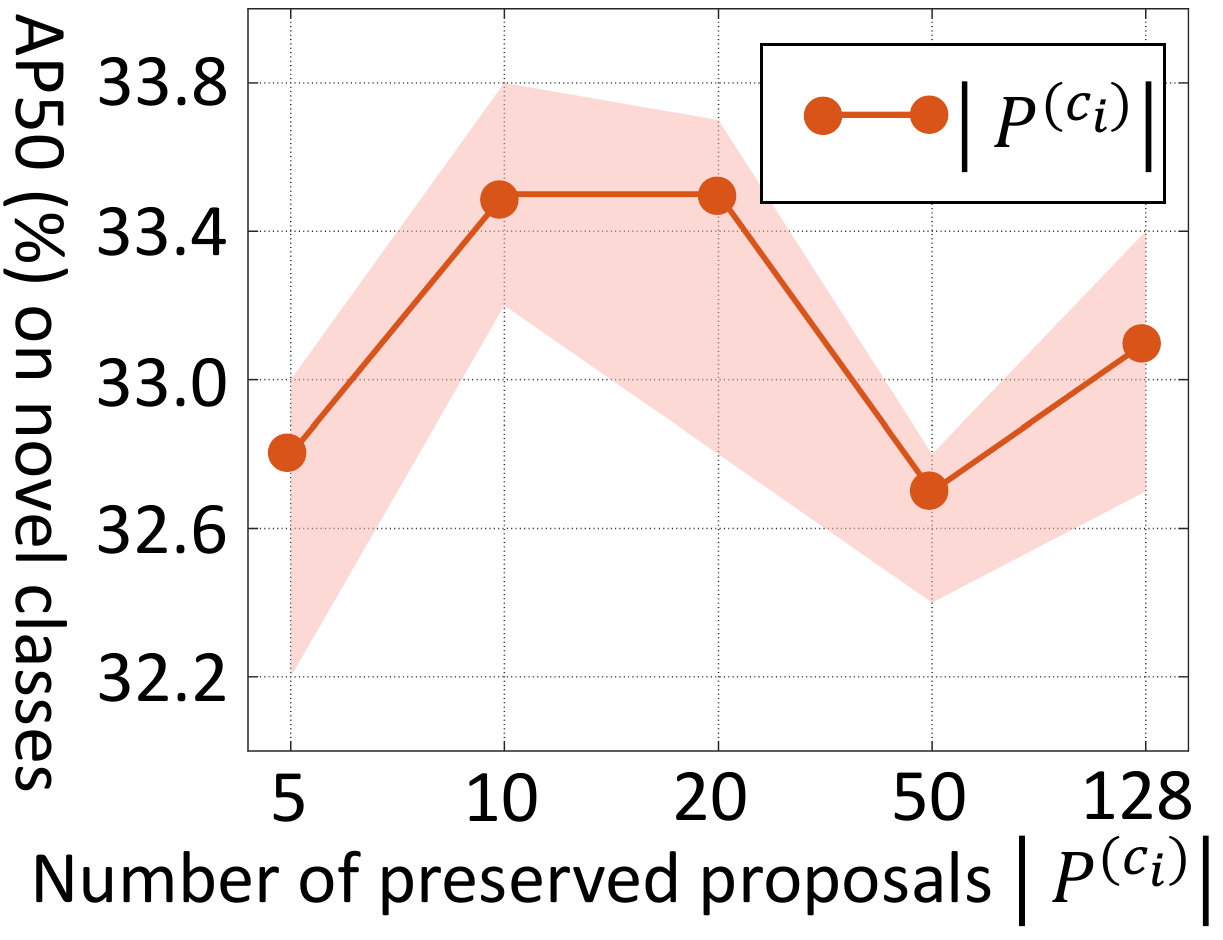}
\caption{Effect of the number of preserved region proposals during pre-filtering. The models are trained and evaluated on COCO.}
\label{fig:p}
\end{figure}

\textbf{Number of preserved proposals in pre-filtering.} We coarsely pre-filter the region proposals before feeding them into the fusion transformer in Eqn.\,(\ref{eqn:candidate_prop}). The number of preserved proposals $|P^{(c_i)}|$ for each concept is ablated in Fig.\,\ref{fig:p}. 
Generally, our approach is relatively robust to the hyper-parameter $|P^{(c_{i})}|$.
To our surprise, MMC-Det still maintains high performance by setting $|P^{(c_{i})}|=128$, which stands for training without the pre-filtering strategy. This indicates that our approach is able to converge to discriminative regions for each concept from massive noise.

\begin{table}[t]
\centering
\caption{Accuracy of MLM predictions on the COCO Captions validation set. Only the object concepts are masked and predicted.}
\begin{tabular}{l|c}
\toprule
Method              & Accuracy ($\%$) \\
\midrule
Vanilla MLM          &  81.7  \\
D-MLM              &  \textbf{85.4}  \\
\bottomrule
\end{tabular}
\label{tab:acc_mlm}
\end{table}

\textbf{Precision of D-MLM}.
To further verify the effectness of contextual knowledge learning of our D-MLM, we evaluate the prediction precision of different masked language modeling approaches. We train two fusion transformers respectively via vanilla MLM and our D-MLM on the COCO Captions training set, and evaluate the masked concept prediction accuracy on the COCO Captions validation set. The results are shown in Tab.\,\ref{tab:acc_mlm}. The results indicate that our D-MLM demonstrates better masked concept prediction accuracy over vanilla MLM, indicating that the region-level information helps masked language prediction.
Meanwhile, high masked concept prediction accuracy ensures the robustness of our noise removal strategy. During our multi-modal contextual knowledge distillation, we recognize the mismatched concepts as noise concepts and exclude them from the calculation of our attention-based contrastive loss in Eqn.\,(\ref{eqn:loss_contr_attn}) and Eqn.\,(\ref{eqn:loss_align}). Too many mismatched concepts in a training batch may shrink the category vocabulary and cause performance degradation. Our D-MLM shows high accuracy ($85.4\%$), thus maintaining enough concepts to enlarge the category vocabulary.

\begin{table}[t]
   \centering
    \caption{OVD results on LVIS dataset~\cite{lvis} with different text encoder. Additionally, we show the zero-shot results that solely trained on the base classes of detection data without any image-caption data.}
    \label{tab:clipvsbert}
    \resizebox{0.95\linewidth}{!}{
    \begin{tabular}{l|cccc}
        \toprule
        Method & $\text{AP}_{\text{r}}$ & \color{gray}{$\text{AP}_{\text{c}}$} &  \color{gray}{$\text{AP}_{\text{f}}$} &\color{gray}{$\text{AP}$} \\
        \midrule
        Zero-shot (BERT~\cite{bert}) &  4.1 &  \g{30.0}  &  \g{35.0} &\color{gray}{27.1}  \\
        Zero-shot (CLIP~\cite{clip}) &  16.3 &  \g{31.0}  &  \g{35.4} &\color{gray}{30.0}  \\
        \midrule
        MMC-Det + BERT~\cite{bert} & 16.4  &   \g{\textbf{31.3}} & \g{35.0} &\color{gray}{30.2}  \\
        MMC-Det + CLIP~\cite{clip} &  \textbf{21.1}  & \g{30.9}  & \g{\textbf{35.5}} &\color{gray}{\textbf{31.0}} \\
        \bottomrule
    \end{tabular}
    }
\end{table}

\textbf{Scalability of text encoder.} 
During evaluating on the LVIS~\cite{lvis} dataset, we empirically observe that the BERT encoder, that performs well on COCO~\cite{coco}, demonstrates low scalability on larger categories (\emph{i.e.}, 1,203 classes in LVIS).
We conduct two experiments to investigate the difference of the two text encoders. (1) Zero-shot setting: the models are trained on the base classes of the detection data and directly evaluated on the testing datasets. (2) Open-vocaulary setting: the models are trained on both the base classes of the detection data and the image-caption data.
As illustrated in Tab.~\ref{tab:clipvsbert}, the models trained with BERT show dramatically performance degradation on novel classes (rare classes) compared to the ones trained with CLIP on both settings.
The reason lies in the fact that BERT lacks visual information, since it is pretrained with pure texts, while CLIP takes both texts and images into consideration. The cross-modal transferability of CLIP ensures its scalability on large visual categories.



\section{Discussions}
\textbf{Conclusions.}
In this paper, we explore multi-modal contextual knowledge for open-vocabulary object detection via deploying a multi-modal contextual knowledge distillation framework called MMC-Det.
This distillation framework transfers the multi-modal contextual knowledge learned in a teacher fusion transformer with diverse multi-modal masked language modeling (D-MLM) to a student detector. The D-MLM serves to learn fine-grained region-level contextual knowledge in the teacher fusion transformer.
Through extensive experiments, we verify that  the contextual knowledge modeled by MMC-Det well benefits OVD.
We hope our study could inspire more new perspectives for open-vocabulary detection.

\textbf{Limitations.}
Despite the impressive performance, two limitations exist for future improvement. 
The first is underutilization of hard positive concepts. Our approach removes the mismatched concepts via D-MLM prediction, which may remove those concepts that are indeed in the images but mis-predicted by the D-MLM. 
The second pertains to computational cost. Although removed during inference, the heavy fusion transformer accounts for non-negligible computational costs during training. Therefore, future improvements may include more efficient training.

\textbf{Broader impact.}
Our research for the first time explores multi-modal contextual knowledge for open-vocabulary detection~(OVD), instead of simple concept-region dot-product similarity matching. Therefore, we provide a new perspective for OVD.
Since the multi-modal supervision is unique to OVD compared to vanilla detection tasks, we hope our study could inspire more perspectives to reasonably utilize the inner high-level semantic information in both the language and vision modalities for open-vocabulary vision tasks.





{
\bibliographystyle{IEEEtran}
\bibliography{egbib}
}

\end{document}